\documentclass[letter]{ieice}
\usepackage[dvipdfmx]{graphicx,xcolor}
\usepackage[fleqn]{amsmath}
\usepackage{algorithm,algorithmic} 
\usepackage{amsmath}  
\usepackage{bm}
\usepackage{graphicx}
\usepackage{comment}

\newcommand{\jtextd}[1]{}  

\newcommand{\AmSLaTeX}{%
 $\mathcal A$\lower.4ex\hbox{$\!\mathcal M\!$}$\mathcal S$-\LaTeX}
\def\BibTeX{{\rmfamily B\kern-.05em%
 \textsc{i\kern-.025em b}\kern-.08em%
  T\kern-.1667em\lower.7ex\hbox{E}\kern-.125emX}}
\makeatletter
\def\tmpcite#1{\@ifundefined{b@#1}{\textbf{?}}{\csname b@#1\endcsname}}%
\makeatother

\field{D}
\vol{107}
\no{1}
\title[Projection-based  Adversarial Attack
using PITL Optimization
]
{%
Projection-based Adversarial Attack
using Physics-in-the- Loop Optimization
for Monocular Depth Estimation}
\authorlist{%
 \authorentry{Takeru KUSAKABE}{n}{labelA}\MembershipNumber{}
 \authorentry{Yudai HIROSE}{n}{labelA}\MembershipNumber{}
 \authorentry{Mashiho MUKAIDA}{m}{labelA}\MembershipNumber{XXXXXXX}
 \authorentry{Satoshi ONO}{m}{labelA}\MembershipNumber{0531898}
}
\affiliate[labelA]{The author is with the
\EICdepartment{\texttt{%
Graduate School of Science and Engineering}}
\EICorganization{\texttt{Kagoshima University}}~\\
\EICaddress{\texttt{Korimoto 1-21-40, Kagoshima, 890-0065 Japan}}
}
\begin{document}
\maketitle
\begin{summary}
Deep neural networks (DNNs)
remain
vulnerable to adversarial attacks that cause misclassification when
specific perturbations are added to input images.
This vulnerability also threatens the reliability of DNN-based
 monocular depth estimation (MDE)
models, making robustness enhancement a critical need in practical
applications.
To validate the vulnerability of DNN-based MDE models, this study
proposes a projection-based adversarial attack method that projects
perturbation light onto a target object.
The proposed method employs physics-in-the-loop (PITL)
optimization---evaluating candidate solutions in actual environments
to account for device specifications and disturbances---and utilizes a
distributed covariance matrix adaptation evolution strategy.
Experiments confirmed that the proposed method successfully created
adversarial examples that lead to depth misestimations, resulting in
parts of objects disappearing from the target scene.
\end{summary}
\begin{keywords}
Adversarial Examples, Monocular Depth Estimation, Physical Attack, Physics-in-the-loop
\end{keywords}

\section{Introduction}\label{intro}

Monocular depth estimation (MDE), which infers three-dimensional data
from scenes
captured by
a single camera \cite{hoi20b19monocular},
has
achieved
significant performance gains in recent years due to the
rapid progress of deep neural networks (DNNs).
Currently, this technology finds potential applications in automated
material handling in factories and warehouses, as well as in
automobile and drone sensing.
However, researchers have revealed that DNNs remain vulnerable to
adversarial examples (AEs) that induce misclassification through
specific perturbations~\cite{goodfellow2014explaining}, and similar
risks affect DNNs used for
MDE.
When such systems support autonomous
robots or self-driving
vehicles, misestimations may lead to accidents.
Consequently, enhancing
the robustness of
MDE models is imperative.

In general, adversarial attacks divide into white-box attacks, which
use internal information such as model parameters and loss gradients,
and black-box attacks, which do not rely on internal data.
Given that AI-based services and commercial systems often restrict
access to their internal architectures and parameters, black-box
attacks are
practically
valuable for externally assessing system
vulnerabilities.

\begin{figure}[t]
    \centering
    \includegraphics[width=0.42\textwidth]{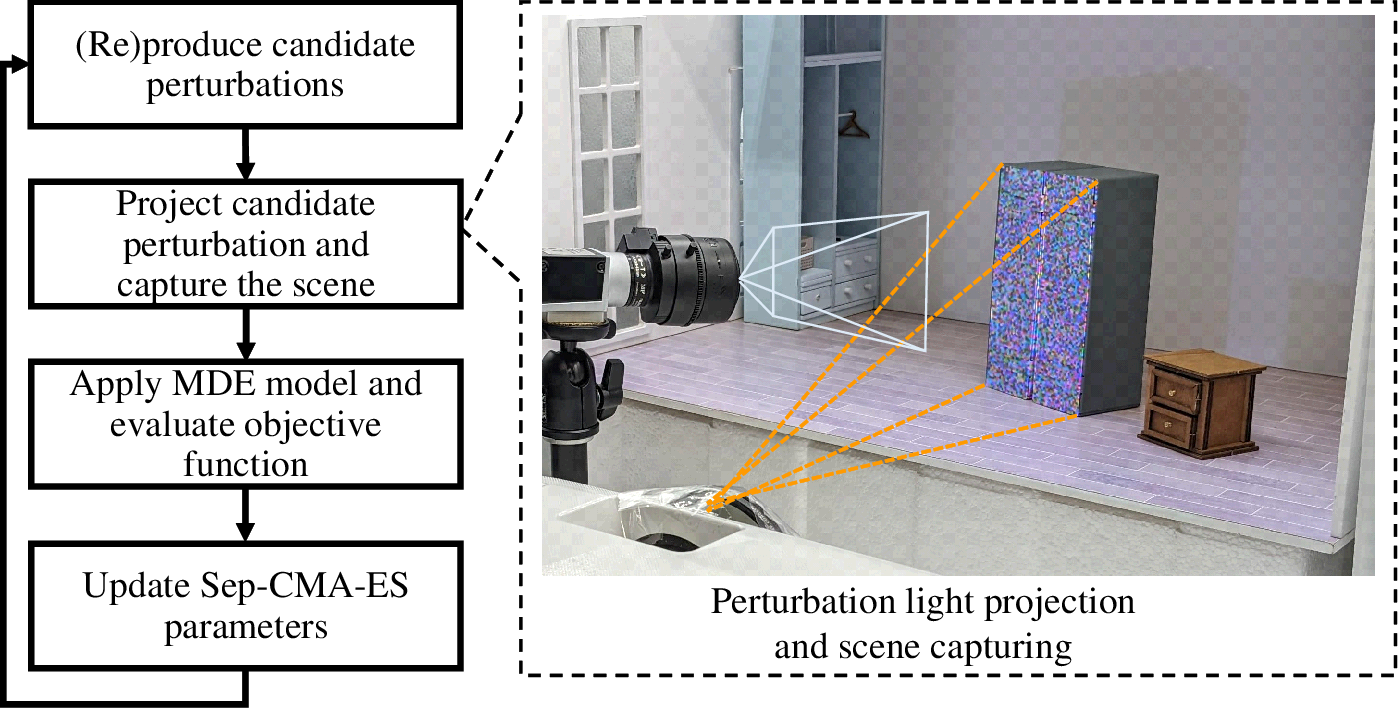}
    \caption{Overview of the proposed method.}
    \label{fig:processing}
\end{figure}

Although early work on adversarial attacks for
DNNs
emphasized digital approaches---such as injecting perturbations or
patches into images~\cite{wang2022survey}---current research
increasingly investigates physical attacks%
on real-world
objects or scenes~\cite{10268441}.
Because these attacks can be performed without compromising the target
system, they more accurately reflect practical threat scenarios.
Identifying physical adversarial examples not only
validates a system's robustness but also quantifies the maximum
disturbances that impair DNN performance.

This paper proposes an adversarial attack method that projects 
perturbation light onto a target object, thereby triggering
misclassifications in MDE systems.
Our method harnesses physics-in-the-loop (PITL)
optimization~\cite{minamata2024coded} to design perturbations by
conducting projection and imaging in real-world settings, effectively
accounting for environmental complexities.
We further integrate
Separable Covariance Matrix Adaptation Evolution Strategy
(sep-CMA-ES)~\cite{ros2008simple}
---%
recognized as one of the
most effective algorithms for high-dimensional black-box optimization---to generate
AEs.
Experiments confirmed that our method induced significant
misestimations that resulted in the disappearance of certain object
regions.

\section{Related work}\label{usage}

Physical adversarial attacks
are
classified into
patch-, camouflage-, and projection-based
methods~\cite{wang2022survey}.
Patch-based and camouflage-based techniques
require direct physical engagement with the target and are thus
invasive, whereas projection-based techniques
are non-ivasive.
The non-invasive category is especially alarming
as
it does not
involve direct physical contact with the object and
causes
perturbations similar to those produced by natural environmental
conditions~\cite{sato2024invisible}.

Daimo et al. proposed a method to generate
AEs
for
MDE models under black-box
conditions~\cite{RenyaDAIMO20232022MUL0001}.
They designed
adversarial projection patterns
via computer graphics
simulation and 
applied
them in a
real-world setting.
However, performance
lagged behind
simulation
due to
difficulty 
in
simulating real-world
factors
such as
reflectance, ambient light, and camera
noise.

\section{The proposed method}\label{generalnote}

\subsection{Key ideas}

This paper proposes a projection-based adversarial attack method for
MDE models.
Compared with patch-based attacks~\cite{cheng2022physical}, a
projection-based attack is limited in the magnitude of perturbation
that can be applied;
it cannot decrease an object's surface brightness, and its
ability to increase brightness is also constrained, leaving the object's
inherent color influence unavoidable.
To mitigate the difficulties associated with projection-based attacks and
address
the issues in previous work~\cite{RenyaDAIMO20232022MUL0001},
our study introduces key ideas outlined below.

\noindent
{\bf Idea 1: Employing PITL optimization.}\\
The proposed adversarial attack method generates perturbation light
patterns by employing
PITL
optimization, which
evaluates candidate solutions under real-world
conditions~\cite{minamata2024coded}.
This method produces adversarial examples that effectively mislead MDE
models in real-world environments, eliminating the need to 
develop
simulators that incorporates complex factors like object reflectance
and ambient light.

\noindent
{\bf Idea 2: Black-box adversarial attack using Separable Covariance
  Matrix Adaptation Evolution Strategy (sep-CMA-ES)}\\
Our proposed method employs sep-CMA-ES
to achieve global optimization without using the
gradient of the objective function, enabling a black-box attack
requiring depth maps produced by a victim MDE model.
This approach facilitates vulnerability analysis for commercial
models, systems, and devices whose internal details are difficult to
access.
Furthermore, sep-CMA-ES
allows us to overcome the limitations of the
previous method~\cite{RenyaDAIMO20232022MUL0001}, which relied
on
combinations of a few local patterns for perturbation generation; the
proposed method facilitates the versatile design of perturbations
projected onto
a target object.

\begin{figure}[t]
  \centering
  \begin{algorithm}[H]
      \caption{The proposed attack algorithm}
      {\footnotesize
          \label{alg:whole}
          \begin{algorithmic}[1]

              \REQUIRE 
                      target depth map
                      $\bm{d}^{(tgt)}$,
                      victim MDE model $h$
              \ENSURE designed perturbation light $\bm{\delta}^*$
              \STATE $g \leftarrow 1$
              \STATE Initialize sep-CMA-ES parameters $\bm{m}^{(1)}$, $\sigma^{(1)}$ and $\bm{C}^{(1)}$
              \WHILE{ $g \leq g_{max}$}
              \FOR {$k \in \{1, \ldots, \lambda\}$}
              \STATE Create perturbation, i.e.,
                     $\bm{\delta}^g_k \leftarrow
                      \mathcal{N}(\bm{m}^{(g)}, {\sigma^{(g)}}^{2}\bm{C}^{(g)})$
              \STATE Project $\bm{\delta}^g_k$ to a target object
              \STATE Capture a scene image $\bm{i}^g_k$
              \STATE Apply MDE model $h$ to $\bm{i}^g_k$, i.e.,
                     $\bm{d}^{(est,g)}_k \leftarrow h(\bm{i}^g_k)$
              \STATE Calculate $f(\bm{\delta}^g_k)=\|\bm{d}^{(est,g)}_k-\bm{d}^{(tgt)}\|_1  $
              \ENDFOR
              \STATE $k_{best} \leftarrow {\rm argmin}_k f(\bm{\delta}^{(g)}_k)$
              \IF{$ f(\bm{\delta}^*) > f(\bm{\delta}^{(g)}_{k_{best}})$}
                  \STATE $\bm{\delta}^* \leftarrow \bm{\delta}^{(g)}_{k_{best}}$
              \ENDIF
              \STATE Update $\bm{m}^{(g)}$, $\sigma^{(g)}$, and $\bm{C}^{(g)}$
                     according to \cite{ros2008simple}
              \STATE $g \leftarrow g + 1 $
              \ENDWHILE
              \RETURN $\bm{\delta}^*$
          \end{algorithmic}}
  \end{algorithm}
  \vspace*{-2mm}
  \caption{The proposed algorithm including PITL optimization.}
  \label{sec:processing}
\end{figure}

\subsection{Formulation}

The proposed method generates a light-projection perturbation pattern
$\bm{\delta}$ on a target object surface, represented 
as
RGB values;
thus, while it can brighten the surface, it cannot
darken it.
We define the objective function
as follows:
\begin{equation}
  \label{fun:objectiveFunction}
  \text{minimize}~  f(\bm{\delta}) =
  \sum_{(w,h) \in \bm{R}} \left| d_{w,h}^{(est)}(\boldsymbol{\bm{\delta}}) - d_{w,h}^{(tgt)} \right|
\end{equation}
where $\bm{R}$ denotes a $w\times h$ region on the target
object surface, $d_{w,h}^{(tgt)}$ represents the depth at pixel $(w, h)$ in
the target depth map, and
$d_{w,h}^{(est)}(\bm{\delta})$
represents the depth
in an estimated depth map when
$\bm{\delta}$ is projected.

\subsection{Algorithm}\label{processing}

Fig.~\ref{sec:processing} shows the proposed PITL optimization algorithm using sep-CMA-ES.
This method searches for the optimal perturbation light
$\bm{\delta}^*$ that causes the MDE model $h$ to misestimate
the scene depth so that the output resembles a given target depth map
$\bm{d}^{(tgt)}$.

The proposed method initializes parameters of sep-CMA-ES, including
mean vector $\bm{m}$, step size $\sigma$, and covariance
matrix $\bm{C}$.
It subsequently repeats the following steps until reaching the maximum
generation $g_{max}$:
sample perturbations $\bm{\delta}^{(g)}_k$ with $k \in \{1, \ldots, \lambda\}$ from a normal distribution
$\mathcal{N}$ with parameters $\bm{m}^{(g)}$, $\sigma^{(g)}$,
and $\bm{C}^{(g)}$, i.e., $\bm{\delta}^{(g)}_k = \bm{m}^{(g)} +
\sigma^{(g)} \bm{y}^{(g)}_k$
where
$\bm{y}^{(g)}_k \sim
\mathcal{N}(\bm{0}, \bm{C}^{(g)})$;
project perturbation $\bm{\delta}^{(g)}_k$ onto a target object and capture the resulting image
$\bm{i}^g_k$;
input $\bm{i}^g_k$ into the depth estimator $h$ to obtain an
estimated depth map $\bm{d}^{(est,g)}_k$;
calculate the objective function value;
update the best perturbation $\bm{\delta}^*$ and
sep-CMA-ES parameters;
and generate the next perturbations.
Finally, the method outputs the optimal perturbation $\bm{\delta}^*$.

\newlength{\figwidth}
\setlength{\figwidth}{25mm}
\begin{figure}[t]
  \centering
  {\scriptsize
  \begin{tabular}{@{}l@{~}c@{~}c@{~}c@{}} 
      \rotatebox{90}{~~~~~Input}
               & {\includegraphics[width=\figwidth]{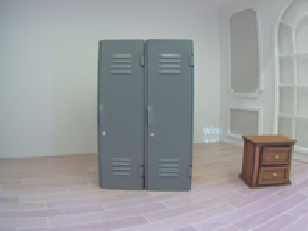}}
               & {\includegraphics[width=\figwidth]{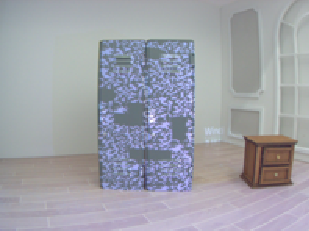}}
               & {\includegraphics[width=\figwidth]{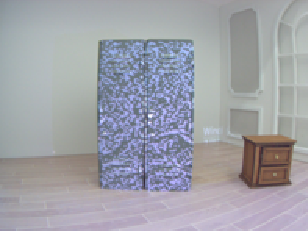}} \\ 

      \rotatebox{90}{~~~Depth map}
               & {\includegraphics[width=\figwidth]{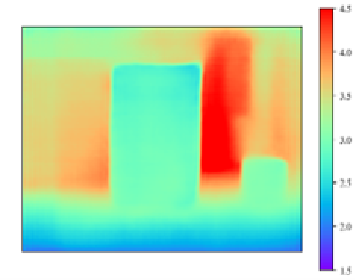}}
               & {\includegraphics[width=\figwidth]{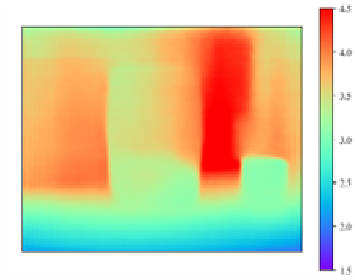}}
               & {\includegraphics[width=\figwidth]{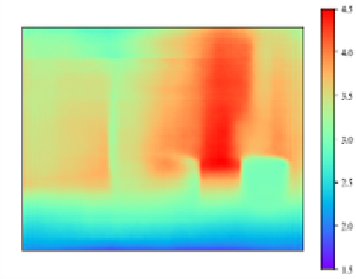}} \\ 
      \rotatebox{90}{~~Point cloud}
               & {\includegraphics[width=\figwidth]{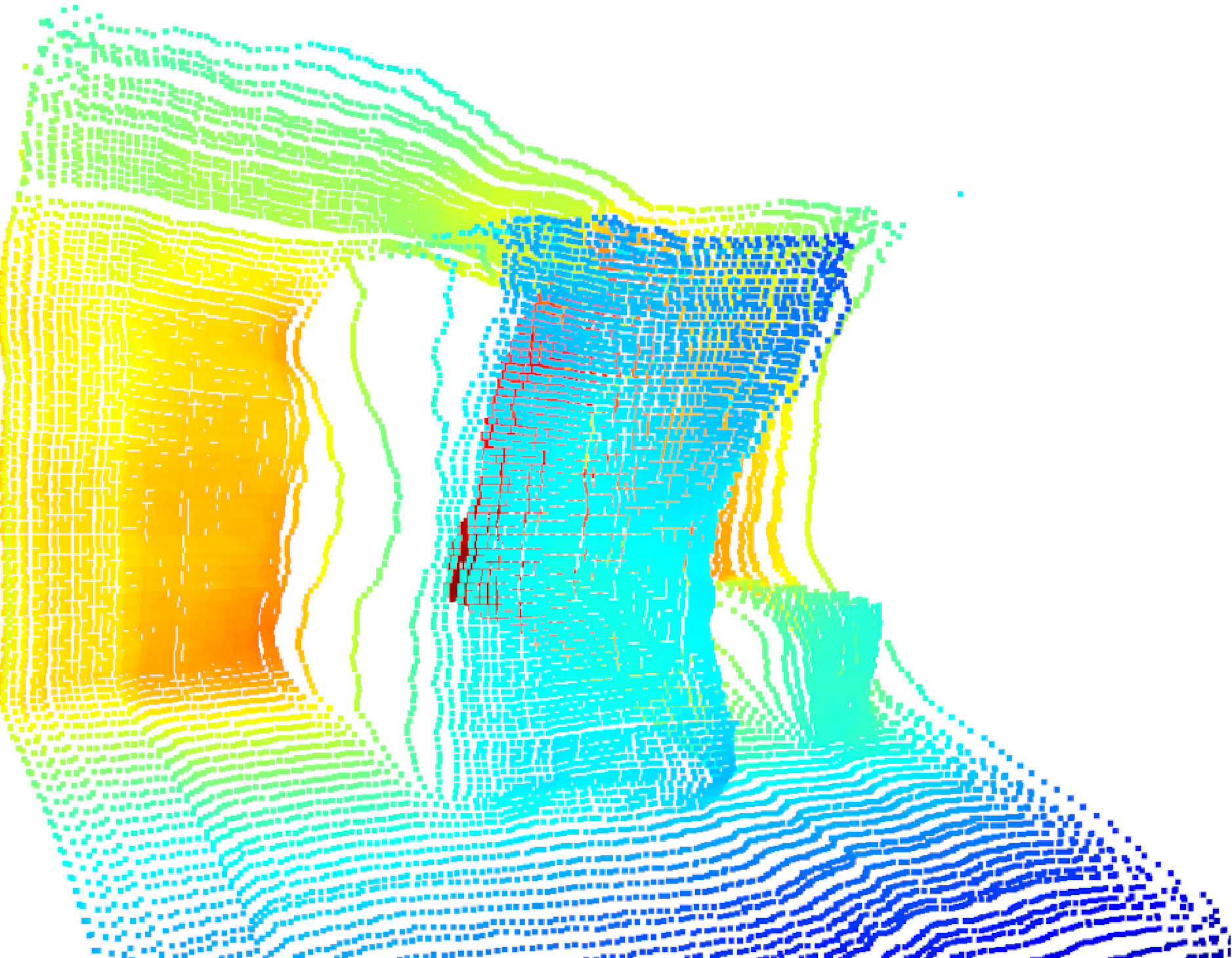}}
               & {\includegraphics[width=\figwidth]{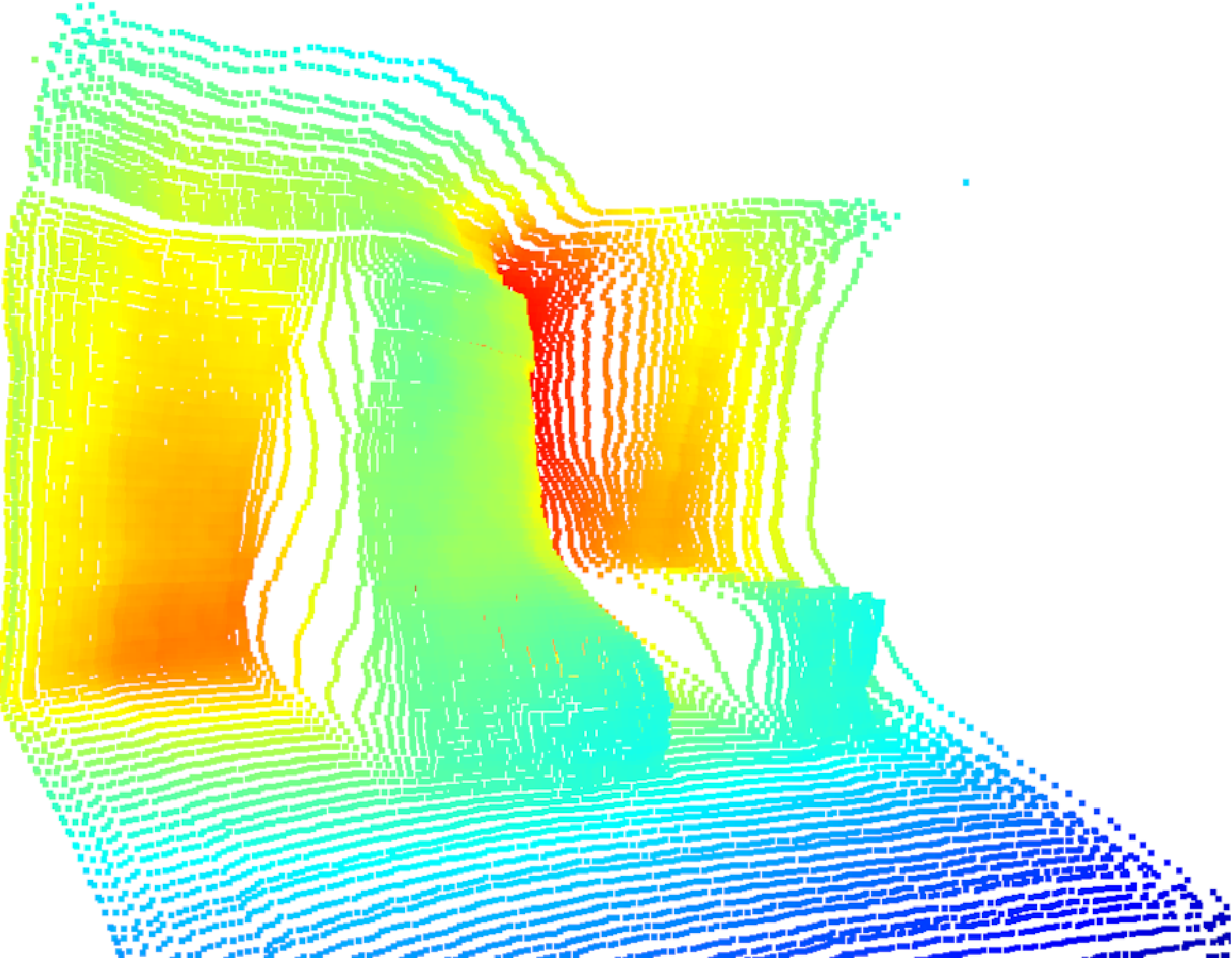}}
               & {\includegraphics[width=\figwidth]{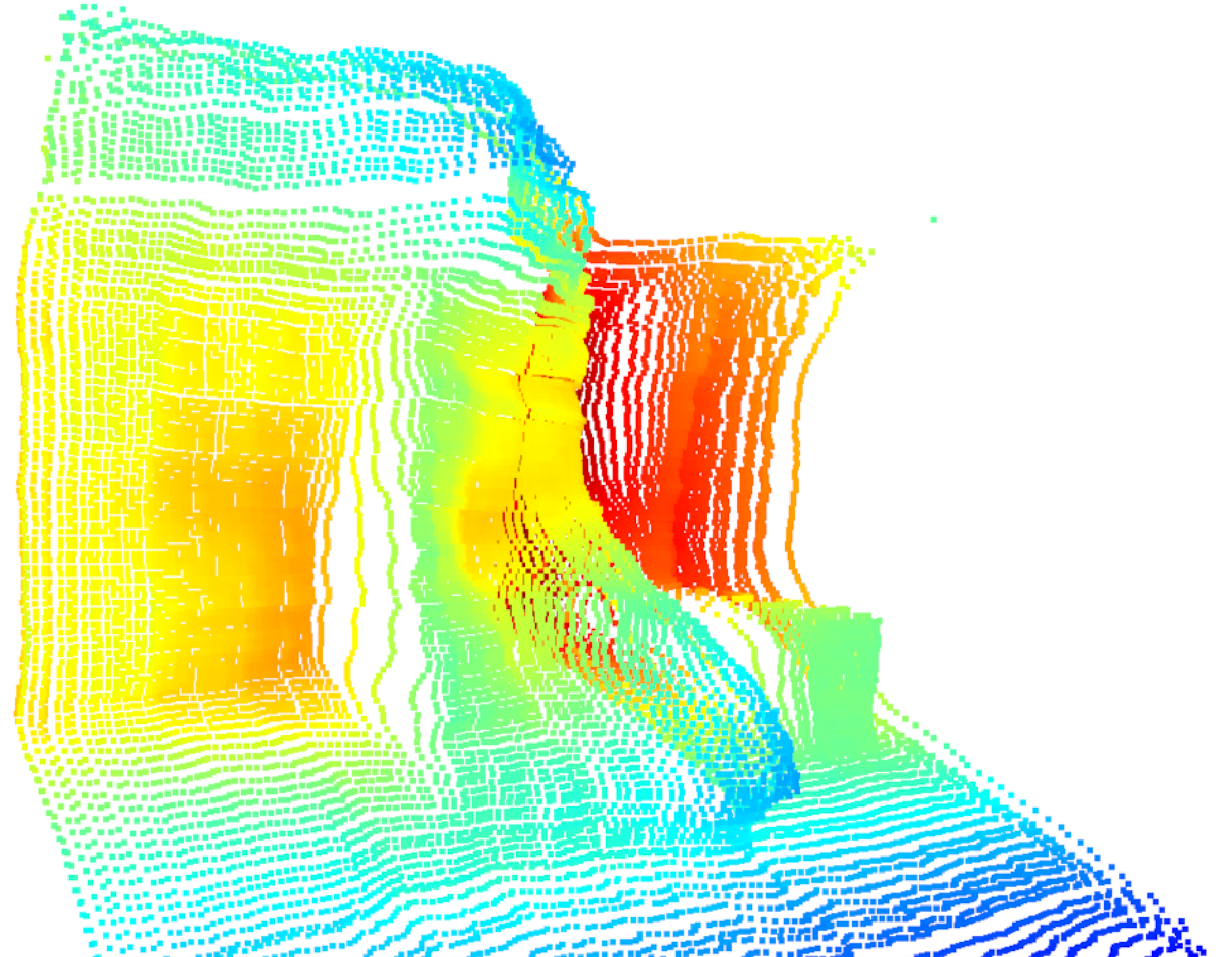}} \\ 
                          & Benign Scenario     & AE by MOEA/D   & AE by CMA-ES \\ 
                          &  ($e=1.000$)
                          &  ($e=0.7512$)
                          &  ($e=0.2399$) \\ 
    \end{tabular}}
  \caption{Exp. 1: Comparison between MOEA/D used in previous work~\cite{RenyaDAIMO20232022MUL0001} and sep-CMA-ES used in our method.}
  \label{tab:result1-a}
~\\ ~\\
  \centering
  {\scriptsize
  \begin{tabular}{@{}l@{~}c@{~}c@{~}c@{}} 
      \rotatebox{90}{~~~~~~~Input}
    & {\includegraphics[width=\figwidth]{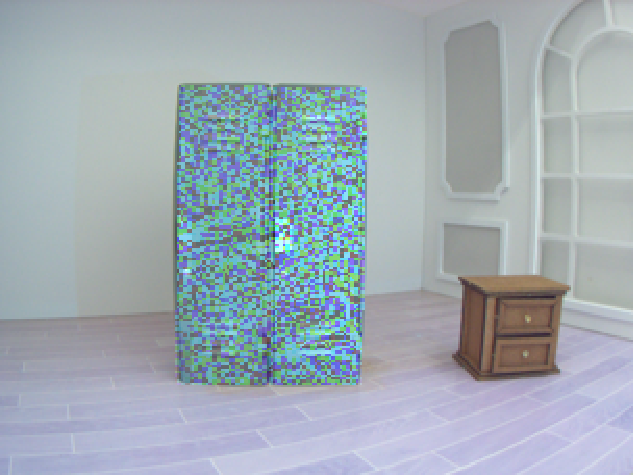}}
    & {\includegraphics[width=\figwidth]{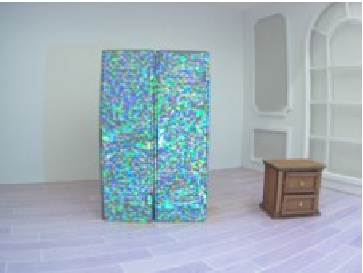}}
    & {\includegraphics[width=\figwidth]{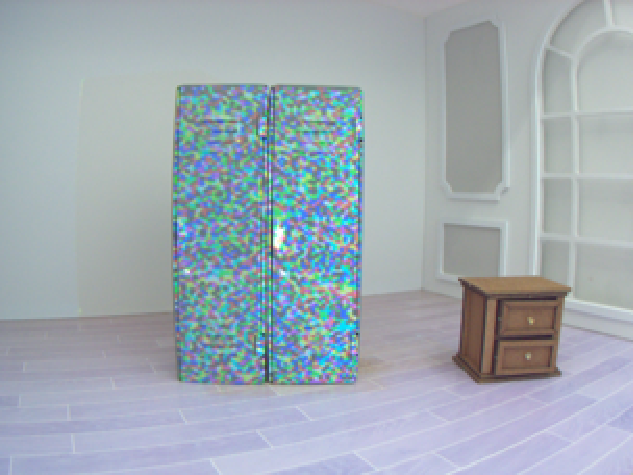}}
    \\

    \rotatebox{90}{~~~Depth map}
    & {\includegraphics[width=\figwidth]{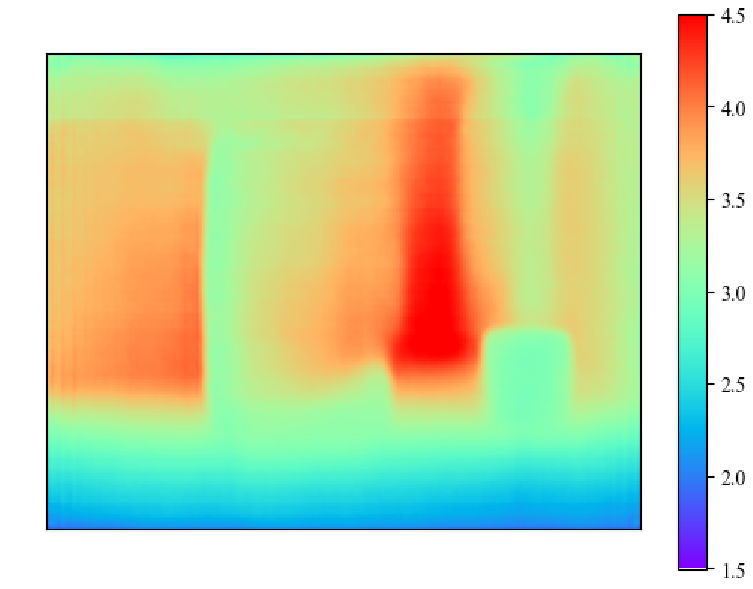}}
    & {\includegraphics[width=\figwidth]{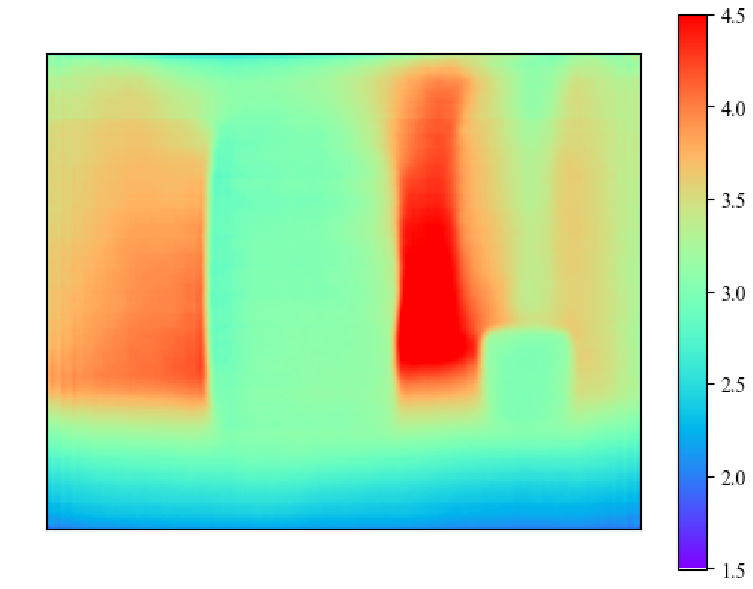}}
    & {\includegraphics[width=\figwidth]{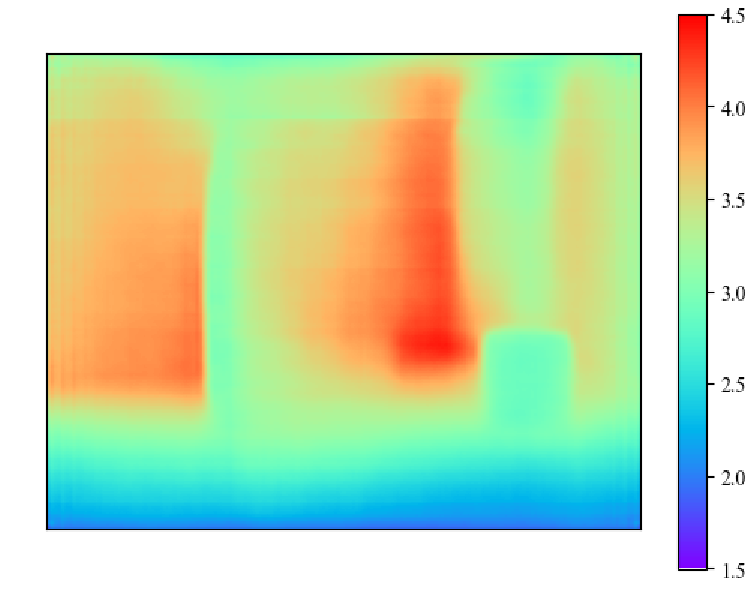}}
    \\
    & \multicolumn{1}{@{~}p{\figwidth}@{~}}{Result in simulation}
    & \multicolumn{1}{@{~}p{\figwidth}@{~}}{Result in real scene}
    & \multicolumn{1}{@{~}p{\figwidth}@{~}}{Result in real scene} \\

    & \multicolumn{2}{@{~}p{2.0\figwidth}@{~}}{AE by sep-CMA-ES using simulation}
    & \multicolumn{1}{@{~}p{\figwidth}@{~}}{AE by our method}\\
    &  ($e=0.2883$)
    &  ($e=0.9303$)
    &  ($e=0.2057$)
    \\
  \end{tabular}}
  \caption{Exp. 2: Comparison with previous method using simulator~\cite{RenyaDAIMO20232022MUL0001}.}
  \label{tab:result_2}
\end{figure}

\section{Evaluation}

To validate our
method, we conducted experiments using a
real
scale model of an indoor scene reproduced at $1/12$ of the actual
size.
In these experiments, we employed a Basler acA1300-30gc camera, a
DV3.4X3.8SA-1 lens, and an EPSON EB-E01 projector.
We defined the target depth map $\bm{d}^{(tgt)}$ as the
image captured 
without the target object
and attempted to generate
AEs
that cause the object to
vanish.
For sep-CMA-ES, $\bm{m}^{(1)}$ was configured so that all RGB values
were set to
their maximum, and $\sigma^{(1)}$ was set to $1.0$.
Population size $\lambda$ was set to $\lambda=4+\lfloor 3 \ln n \rfloor$,
where $n$ denotes the number of variables ($n=3|\bm{R}|$);
thus, $\lambda$ varies slightly across experiments, ranging
from 28 to 32.
Other parameters were
set to
the default values of sep-CMA-ES
~\cite{ros2008simple}.

Moreover, we quantified object presence by defining a presence
rate $e$ as follows:
\begin{equation}
    e = \frac{1}{|\bm{R}|} \sum_{(w,h) \in \bm{R}} 
        \frac{\left| d_{(w,h)}^{(est)} - d_{(w,h)}^{(back)} 
        \right|}{\left| d_{(w,h)}^{(orig)} - d_{(w,h)}^{(back)} \right|}
\end{equation}
where $d_{(w,h)}^{(back)}$ denotes the depth value of the background;
$e$ equals 1 if the object is fully present and 0 if it completely
disappears.

This study validates our  method through four experiments: one
measuring the effect of using sep-CMA-ES (Exp. 1), another
assessing PITL optimization (Exp. 2), a third evaluating
performance on low-height objects (Exp. 3), and a fourth evaluating
a state-of-the-art MDE model (Exp. 4).
Our proposed method performed PITL optimization in real enviroments
across all four experiments.
Except for part of Exp. 2, all evaluations were conducted 
using real scenes.

Exp. 1 compared the sep-CMA-ES optimizer used in our proposed method with
MOEA/D algorithm employed in previous work~\cite{RenyaDAIMO20232022MUL0001},
both of which performed PITL optimization.
We targeted an MDE model by Laina et al.~\cite{laina2016deeper}, trained
on the NYU Depth v2 indoor scene dataset.
Both algorithms were configured with 900 generations and a population
size of 28.

Fig.~\ref{tab:result1-a} shows the results.
Compared with MOEA/D, our method employing sep-CMA-ES produced
perturbations that blurred the boundary in the upper locker area,
leading to more pronounced object disappearance%
\footnote{%
Given the inherent difficulty in completely removing a target object
in depth estimation,
attacks are often judged successful
when they cause even a partial object
loss~\cite{cheng2022physical}.
},
and it also achieved lower $e$ values.
MOEA/D mitigates dimensionality growth through limited local pattern
combinations, but this restricts its perturbation diversity, whereas
the unconstrained sep-CMA-ES produces more significant misestimations.

Exp. 2 assessed the efficacy of PITL optimization in our 
proposed method, i.e., integrating solution candidate evaluation using
actual devices into the optimization loop, compared to
traditional
simulation-based evaluation.
As in Exp. 1, we employed Laina et al.'s model as 
the
victim model.
sep-CMA-ES was configured with
$g_{max}=800$ and
$\lambda=31$.

Fig.~\ref{tab:result_2} illustrates the results:
from left to right, it shows the perturbations and estimated depth
maps of the AE generated by optimization using simulation-based
solution candidate evaluation---assessed in both simulation and real
environments---and those of the AE generated by our method.
The perturbation that, in simulation, led to misestimations causing the
right half of the locker to vanish did not produce such severe
misestimations when projected in the actual environment.
This discrepancy likely results from differences between simulation and
real conditions, yet our PITL approach manages to generate the AE in
real settings that is
comparable to
that from simulation-based
optimization.

Exp. 3 investigated the impact of low-height target objects,
which a previous study~\cite{RenyaDAIMO20232022MUL0001} found challenging
to misestimate in depth, by employing a kitchen gas stove and
a sofa.
Consistent with Exp. 1 and 2, we employed 
Laina
et al. model,
configuring sep-CMA-ES with
$\lambda=30$ and
$g_{max}=800$.

Fig.~\ref{tab:result_3} illustrates the results, which reveal that the
attack on the gas stove induced significant misestimations,
characterized by a blurred object-background boundary and the
vanishing of the left half.
In contrast, the attack on the sofa
caused
nearly complete
disappearance of the object, excluding only the armrest that 
was
not
perturbed.
These results indicate that our
method can effectively
attack
low-height objects.

Exp. 4 aimed to demonstrate our 
method's
applicability to
state-of-the-art MDE models by testing on Depth Anything v2 small
model~\cite{yang2025depth}.
sep-CMA-ES parameters were set to
$\lambda=32$ and $g_{max}=1,700$.

Fig.~\ref{tab:result_4} illustrates that, in the absence of projected
perturbations, the model correctly captures the spatial relationship
in which
the target locker is situated in front of a small shelf on its
right.
However, in the adversarial scene, the estimated depth, except at the
locker's edges, erroneously reaches the wall.
These results confirm that our method demonstrates effective attack
capability even on modern MDE models.

\setlength{\figwidth}{25mm}
\begin{figure}
  \centering
  {\scriptsize
  \begin{tabular}{@{}l@{~}c@{~}c@{~}c@{}} 
      \rotatebox{90}{~~~~Benign}
    & {\includegraphics[width=\figwidth]{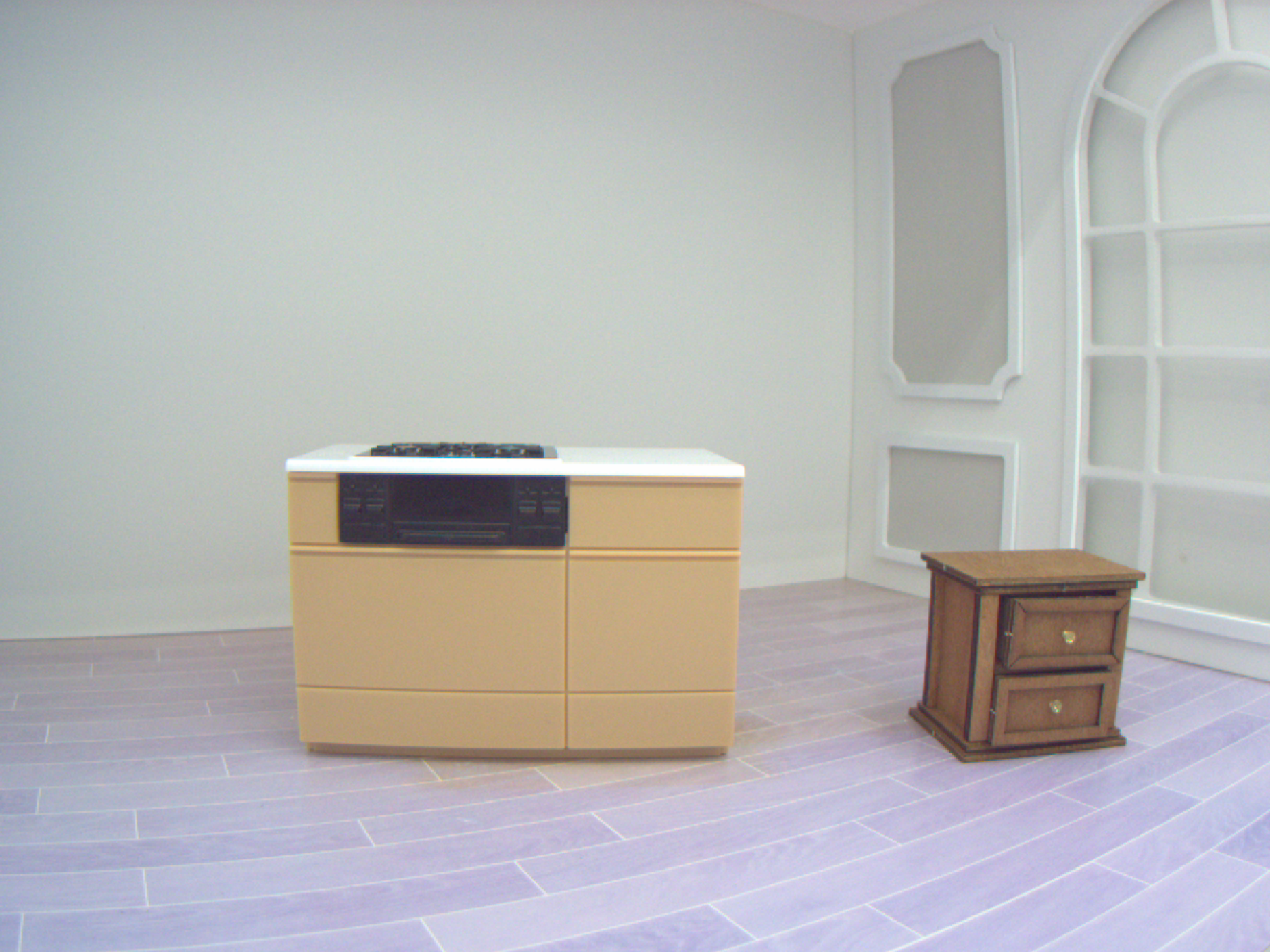}}
    & {\includegraphics[width=\figwidth]{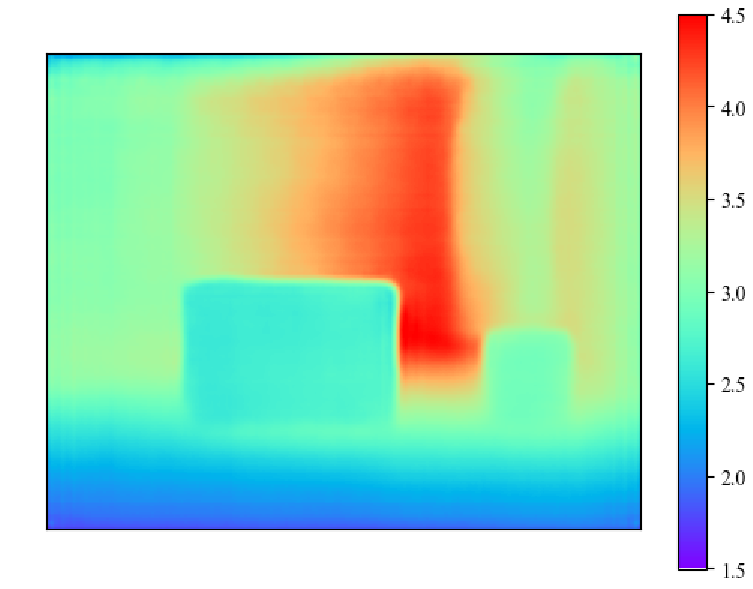}}
      & {\includegraphics[width=\figwidth]{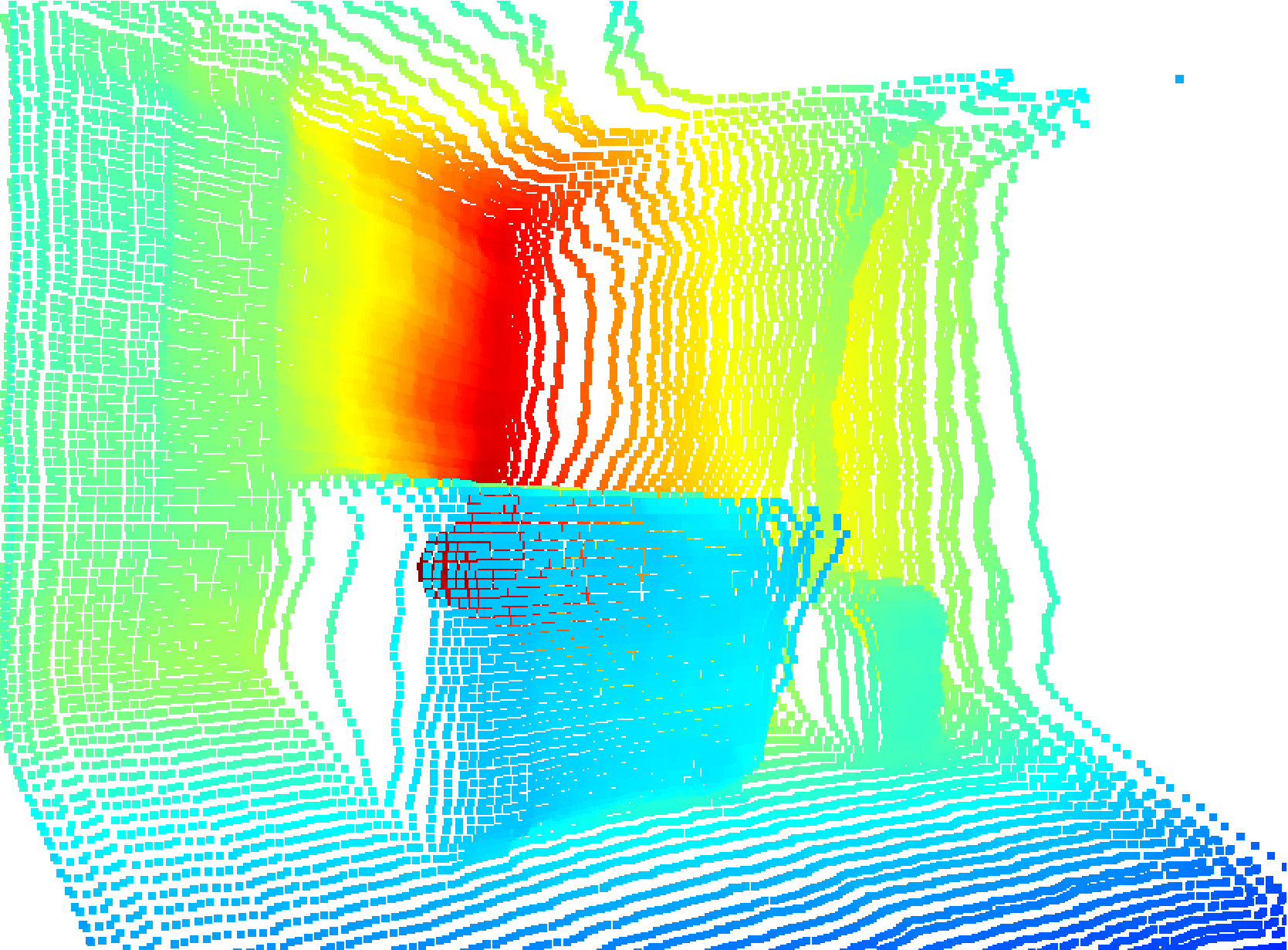}}\\
      \rotatebox{90}{~~~Adversarial}
    & {\includegraphics[width=\figwidth]{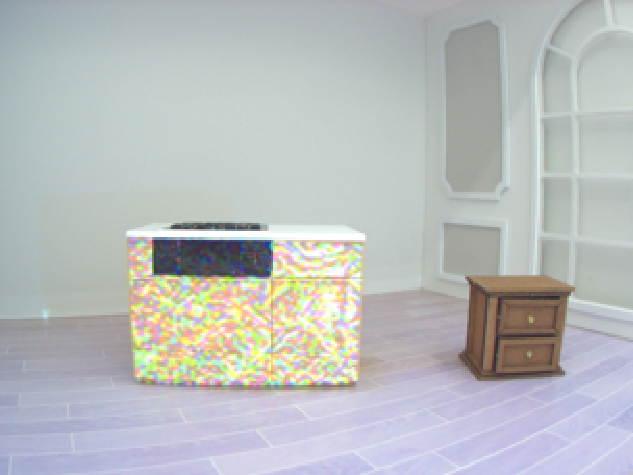}}
    & {\includegraphics[width=\figwidth]{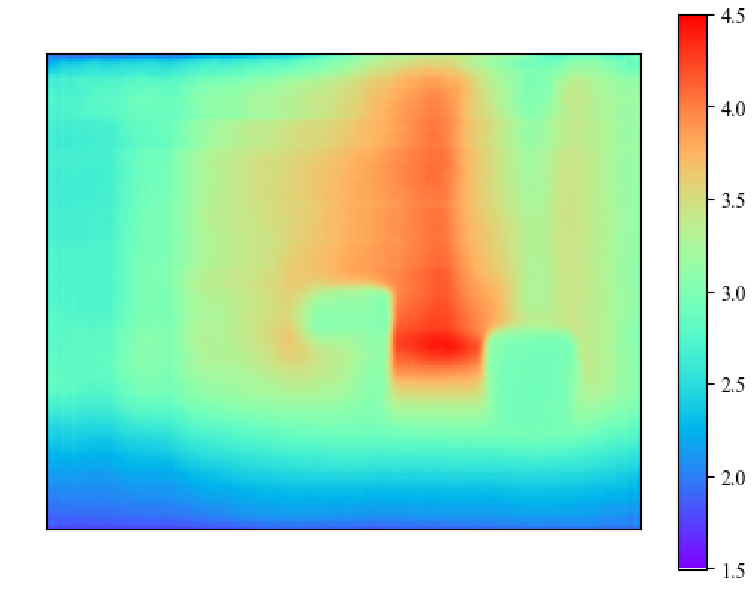}}
    & {\includegraphics[width=\figwidth]{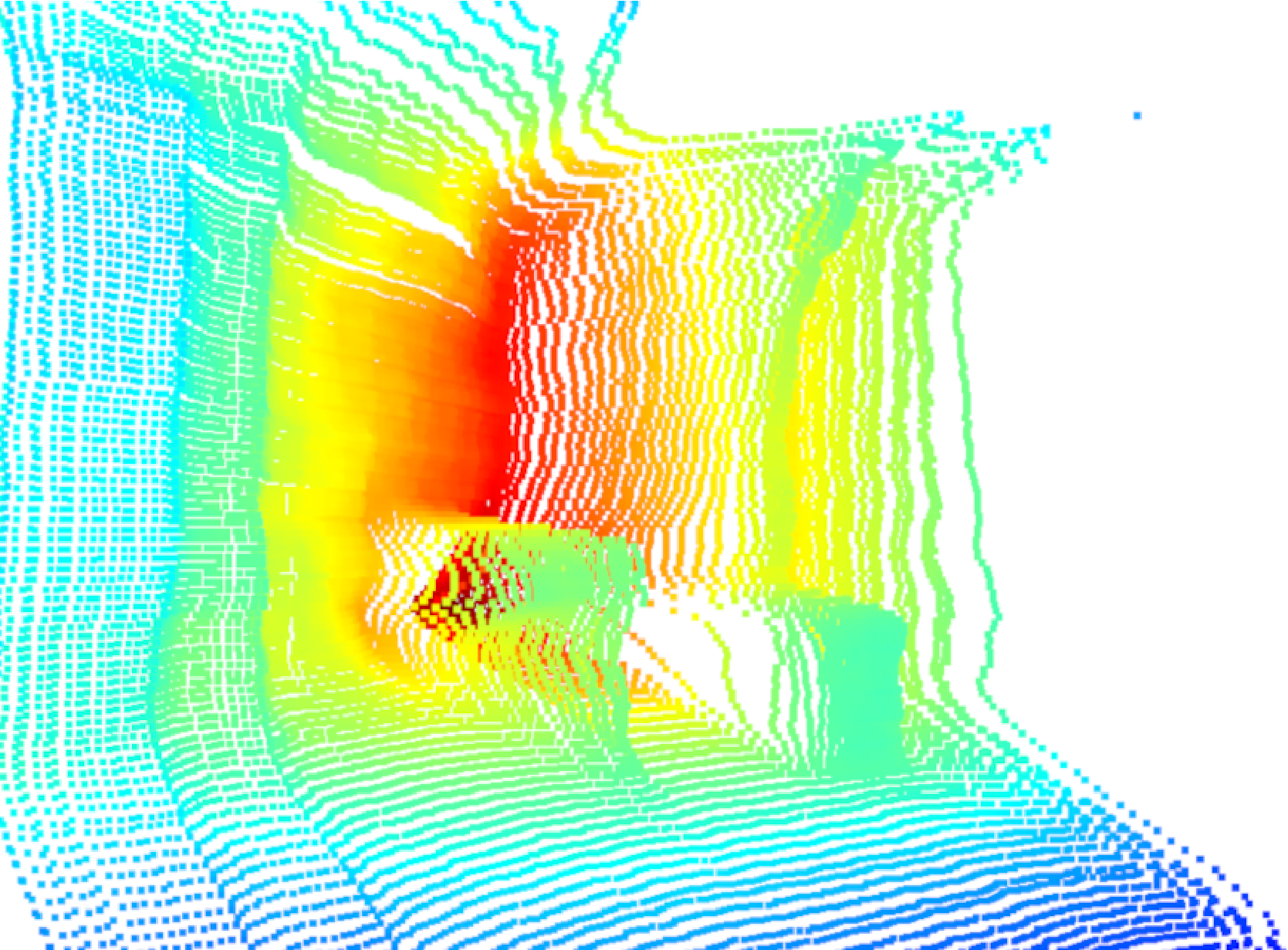}}\\
    & Input & Depth map & Point cloud\\
    \multicolumn{4}{c}{(a) Kitchen gas range ($e = 0.2743$)} \\
      \rotatebox{90}{~~~~Benign}
    & {\includegraphics[width=\figwidth]{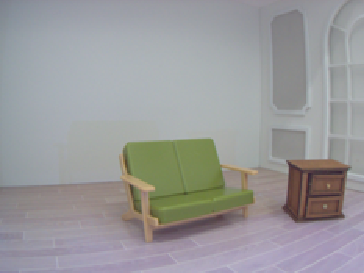}}
    & {\includegraphics[width=\figwidth]{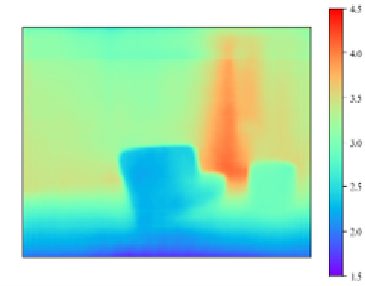}}
    & {\includegraphics[width=\figwidth]{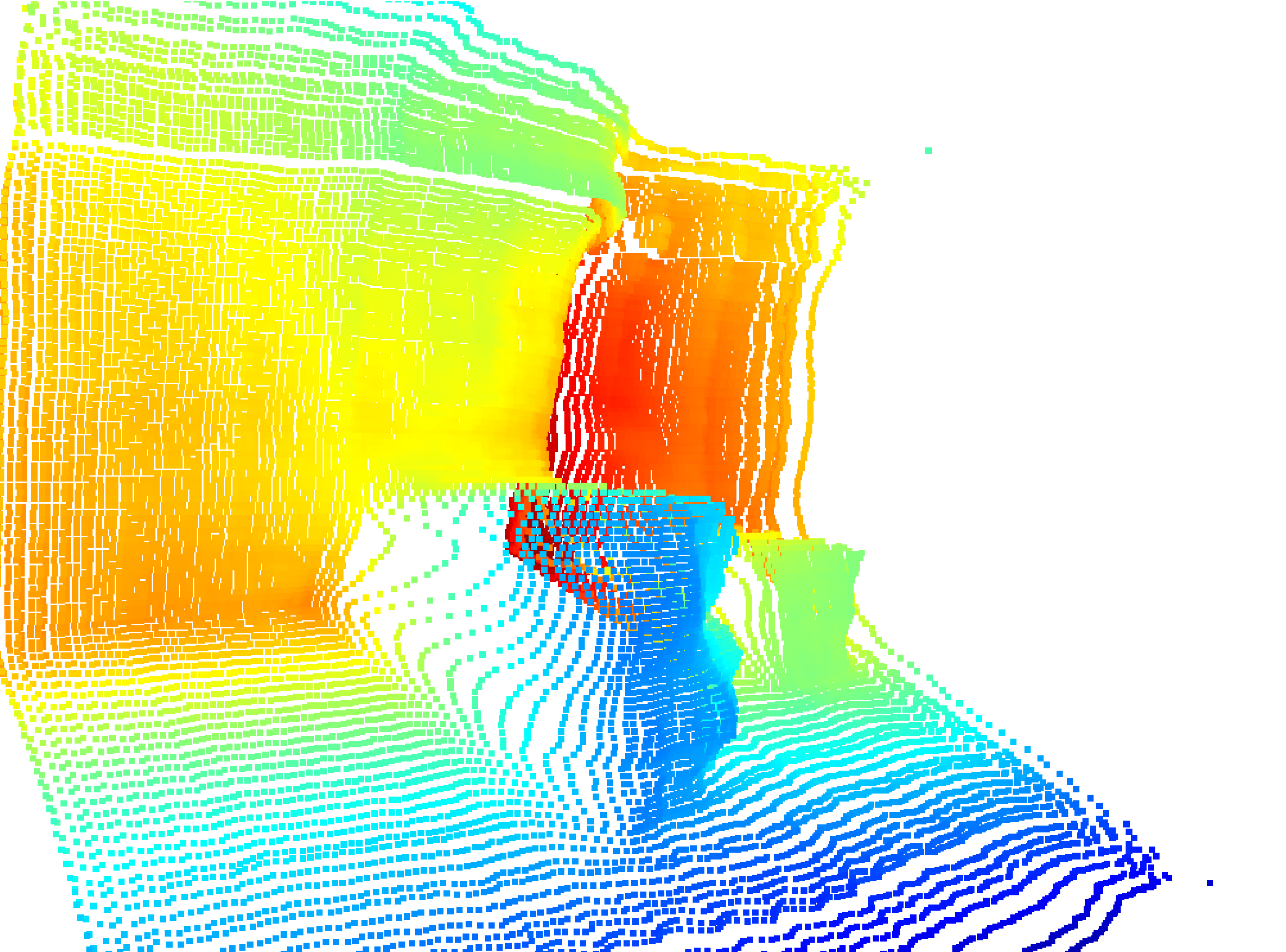}}\\
      \rotatebox{90}{~~~Adversarial}
    & {\includegraphics[width=\figwidth]{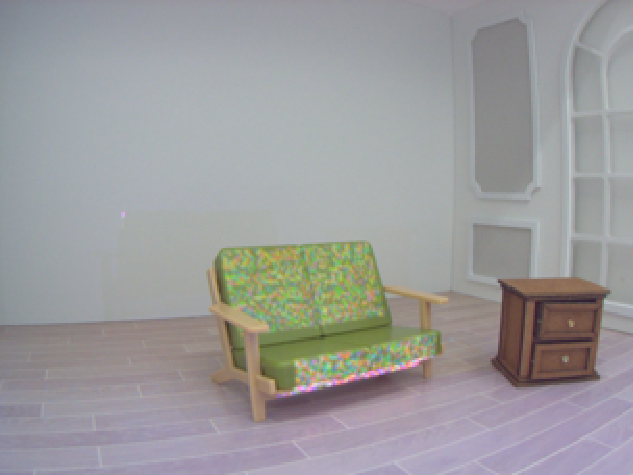}}
    & {\includegraphics[width=\figwidth]{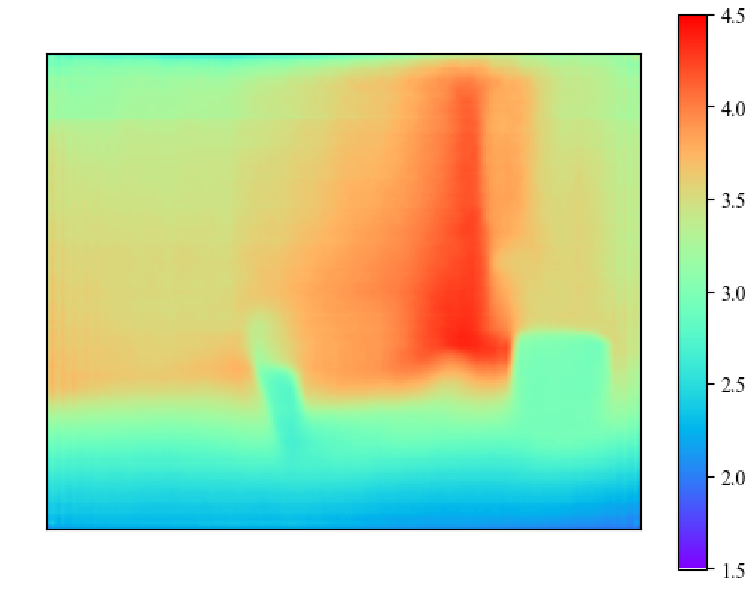}}
    & {\includegraphics[width=\figwidth]{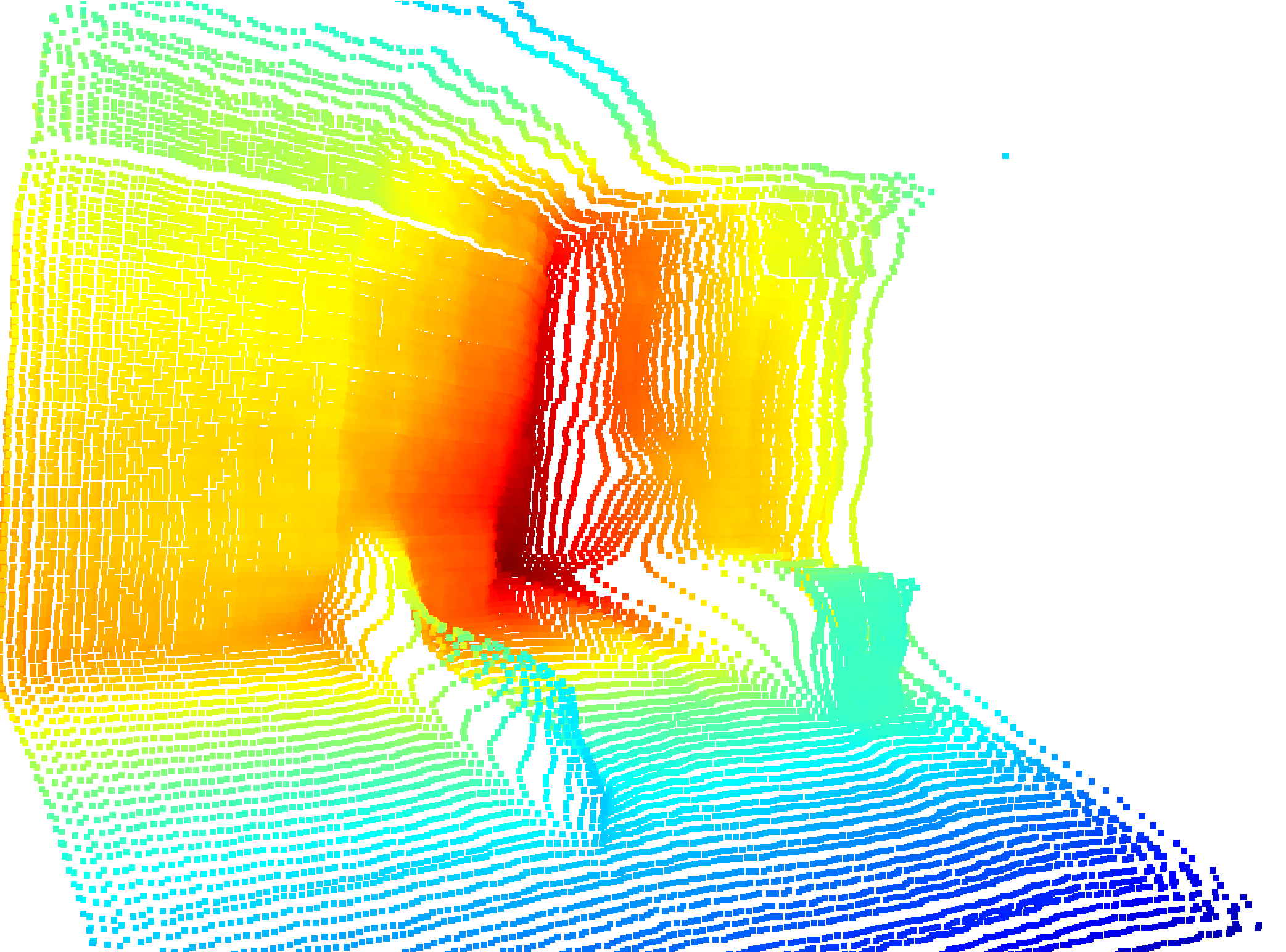}}\\
    & Input & Depth map & Point cloud\\
    \multicolumn{4}{c}{(b) Sofa ($e = 0.0425$)}\\
  \end{tabular}}
  \caption{Exp. 3: results against other objects.}
  \label{tab:result_3}
~\\ ~\\
  \centering
  {\scriptsize
  \begin{tabular}{@{}l@{~}c@{~}c@{~}c@{}} 
      \rotatebox{90}{~~~~Benign}
    & {\includegraphics[width=\figwidth]{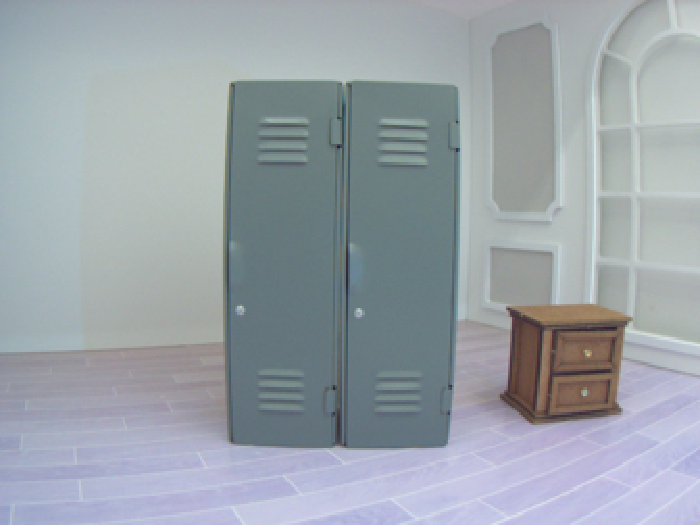}}
    & {\includegraphics[width=30mm]{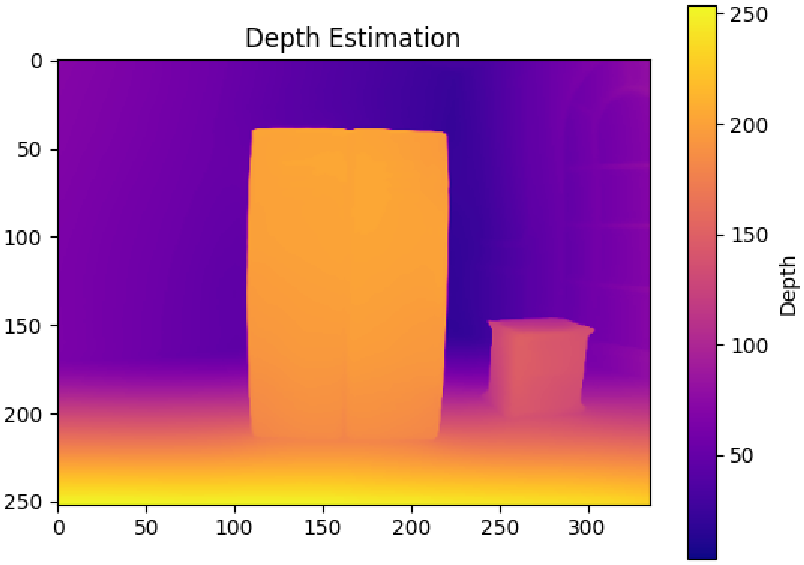}}
    & {\includegraphics[width=26mm]{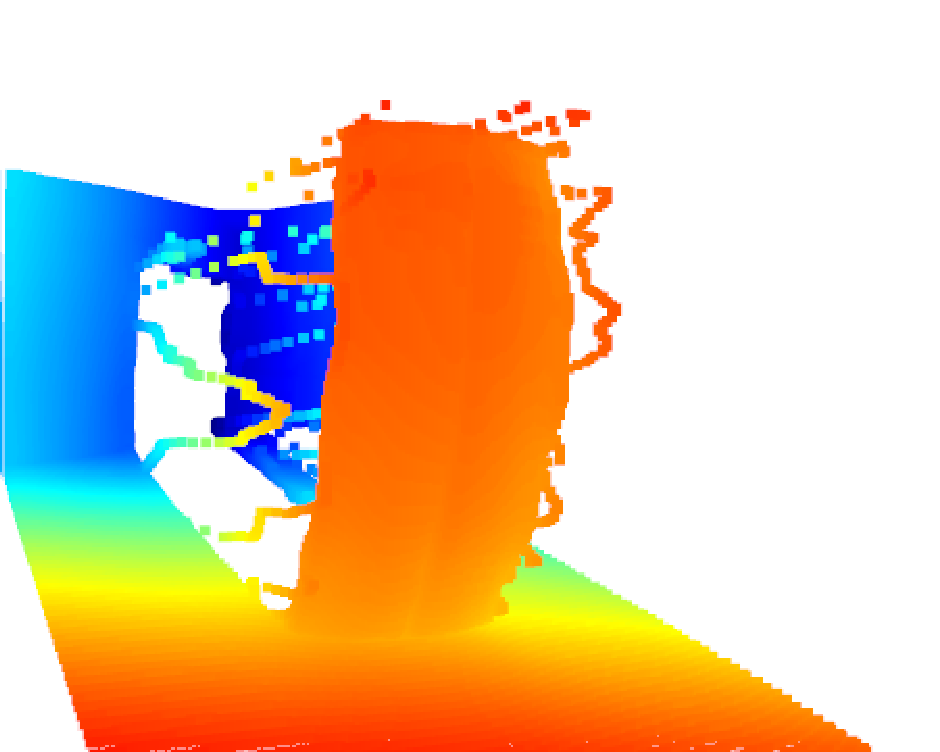}} \\
      \rotatebox{90}{~~Adversarial}
    & {\includegraphics[width=\figwidth]{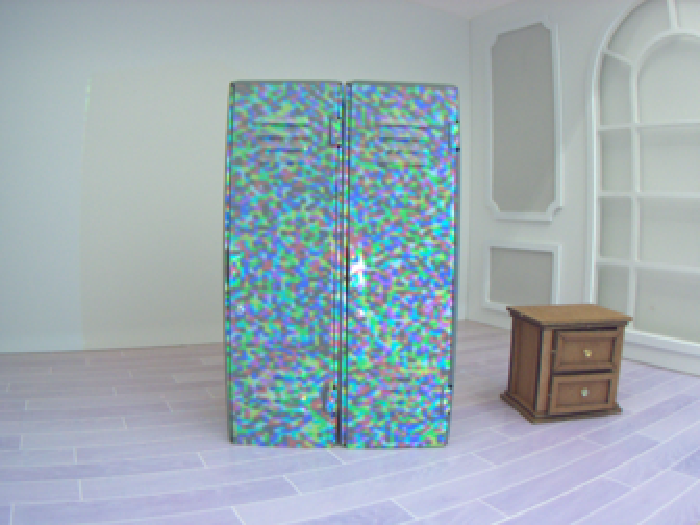}}
    & {\includegraphics[width=30mm]{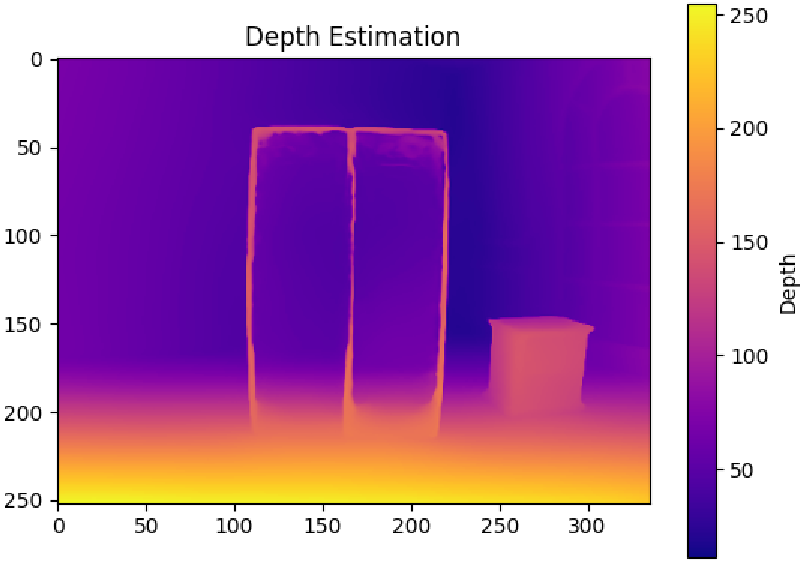}}
    & {\includegraphics[width=26mm]{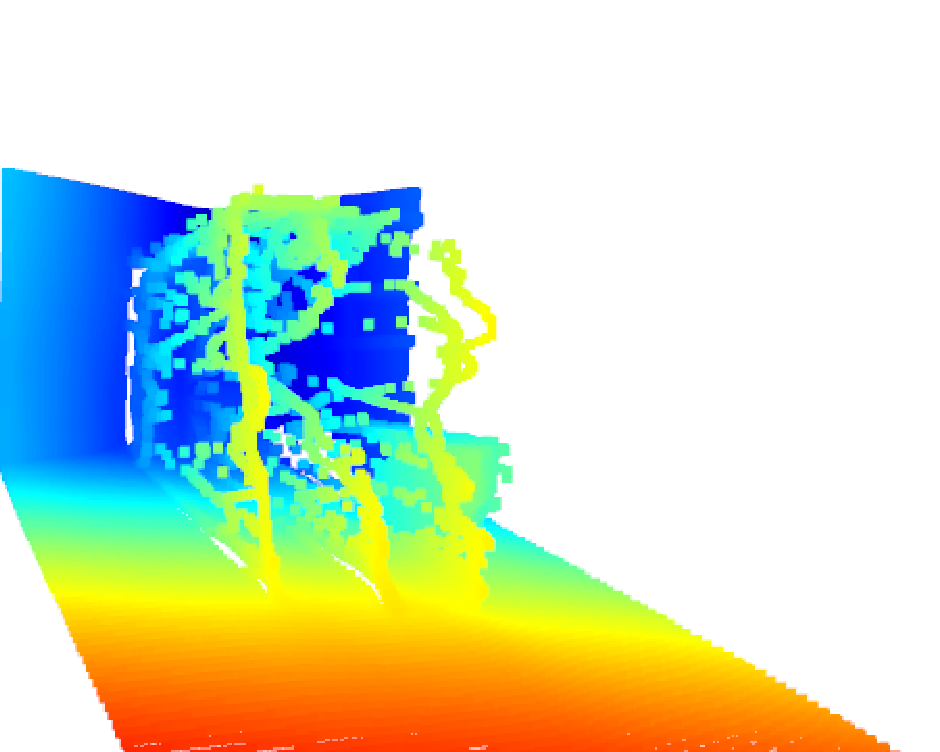}} \\
    & Input & Depth map & Point cloud\\
    \multicolumn{3}{c}{~}\\
  \end{tabular}}
  \caption{Exp. 4: attack to state-of-the-art MDE model Depth Anything v2 ($e=0.312$).}
  \label{tab:result_4}
\end{figure}

\section{Conclusions}

This paper proposed a projection-based adversarial attack method that,
under black-box conditions, generates perturbation light projected onto
a target object surface to attack MDE models.
Our proposed method employed PITL optimization to account for various
physical factors---including projector and camera distortions, ambient
light, and noise---thus enabling the design of effective
perturbation patterns for target scenes.
Moreover, using sep-CMA-ES facilitates the versatile design of
perturbations projected onto an object surface.
Experimental results demonstrated that our method generated
adversarial examples resulting in the disappearance of parts of objects.

On the other hand, a drawback of the proposed method is that its
perturbations are substantial and perceptible. In the future, we
plan to extend our method to reduce the magnitude of perturbations

\bibliographystyle{ieeetr}
\bibliography{myref} 

\end{document}